%% file: main_arxiv.tex
\documentclass{article} 
\usepackage{aditi}

\usepackage{microtype}
\usepackage{hyperref}
\usepackage{url}
\usepackage{booktabs}
\usepackage{titlesec}
\usepackage{enumitem}
\usepackage{amsfonts}
\usepackage{nicefrac}
\usepackage{subcaption}
\usepackage{xcolor}
\usepackage{bm}
\usepackage{graphicx}
\usepackage{amsmath}
\usepackage{amssymb}
\usepackage{amsthm}
\usepackage{xspace}
\usepackage{adjustbox}
\usepackage{tcolorbox}
\usepackage{multirow}
\usepackage{pifont}
\usepackage{caption}
\usepackage{inconsolata}
\usepackage{wrapfig}
\usepackage{float}          
\usepackage[ruled,vlined]{algorithm2e}
\usepackage{tikz}
\usetikzlibrary{arrows.meta,backgrounds}

\usepackage{etoolbox}
\newtoggle{arxiv}
\toggletrue{arxiv}
\usepackage{style}

\definecolor{darkblue}{rgb}{0, 0, 0.5}
\hypersetup{colorlinks=true, citecolor=darkblue, linkcolor=darkblue, urlcolor=darkblue}

\titlespacing*{\subsection}{0pt}{0.5\baselineskip}{0.25\baselineskip}
\titlespacing*{\section}{0pt}{0.5\baselineskip}{0.25\baselineskip}

\usepackage[utf8]{inputenc} 
\usepackage[T1]{fontenc}    

\input{math_commands.tex}

\renewcommand{\cite}[1]{\citep{#1}}
\setcitestyle{authoryear,round,citesep={;},aysep={,},yysep={;}}

\input{specific_macros}

\title{Off-Policy Merging Beats On-Policy Self-Distillation \\ for Continual Learning}

\setauthors{Chen Henry Wu, \authorsep Thomas Zhang, \authorsep Aditi Raghunathan}
\setaffils{Carnegie Mellon University}
\setemail{{\{chenwu2,thomasz4,raditi\}@cs.cmu.edu}}
\setcode{https://github.com/ar-forum/grafting}
\setwebsite{https://ar-forum.github.io/grafting-web/}

\begin{document}

\maketitle

\input{tex/abstract}
\input{tex/intro}
\input{tex/preliminaries}
\input{tex/theory}
\input{tex/method}
\input{tex/related_work}
\input{tex/conclusion}
\input{tex/acks}

\bibliography{main_arxiv}
\bibliographystyle{plainnat}

\appendix
\crefalias{section}{appendix}

\input{appdx/data}
\input{appdx/pedagogical_RL}
\input{appdx/training}
\input{appdx/evaluation}
\input{appdx/olmo3_exps}
\input{appdx/fisher}

\end{document}

%% file: math_commands.tex
\usepackage{amsmath,amsfonts,bm}

\iftoggle{arxiv}{
    
    \def\eqref#1{equation~\ref{#1}}
    
}
{}

\def\1{\bm{1}}

\DeclareMathAlphabet{\mathsfit}{\encodingdefault}{\sfdefault}{m}{sl}
\SetMathAlphabet{\mathsfit}{bold}{\encodingdefault}{\sfdefault}{bx}{n}

\renewcommand{\E}{\mathbb{E}}

\renewcommand{\R}{\mathbb{R}}

\renewcommand{\KL}{D_{\mathrm{KL}}}

\DeclareMathOperator{\Tr}{Tr}

%% file: specific_macros.tex
\newcommand{\lam}{\lambda}
\newcommand{\thpost}{\theta_{\text{post}}}
\newcommand{\thdonor}{\theta_{\text{donor}}}
\newcommand{\thdonorft}{\theta_{\text{donor-ft}}}
\newcommand{\mask}{\mathcal{M}}                                 

\definecolor{donorc}{HTML}{2980b9}   
\definecolor{maskc}{HTML}{27ae60}    
\newcommand{\maskrho}{\textcolor{maskc}{\mask_\rho}}

\iftoggle{arxiv}
{
  \newcommand{\confpar}[1]{\paragraph{#1}}
}
{
  \newcommand{\confpar}[1]{\textbf{#1}}
}

\newcommand{\ip}[1]{\ensuremath{\left\langle #1 \right\rangle}}

%% file: tex/abstract.tex
\begin{abstract}

A long-standing goal of AI is a model that can continually learn and improve itself. On post-trained models, supervised finetuning (SFT) on new data often causes poor generalization and catastrophic forgetting. As such, the conventional wisdom is that \textit{on-policy} training is a prerequisite for continual learning. In practice, however, data containing new knowledge or capabilities are often \textit{off-policy}. While methods such as on-policy self-distillation (OPSD) try to bridge this gap by converting off-policy data into on-policy signal, they have been shown to cause reasoning collapse. 
In this paper, we show that \textit{off-policy merging beats OPSD for continual learning}. 
We first show that SFT learns a useful signal from new data, but naively applying its update interferes with existing capabilities. We reduce this interference with a simple recipe we term \textbf{grafting}, which changes where the update is learned and how it is applied: (1) learning the update on an earlier \textbf{donor checkpoint}, ideally even before the end of \textit{pretraining}, and applying the weight update to the post-trained model; (2) \textbf{scaling} the weight update, equivalent to a form of model merging; and (3) optionally, \textbf{masking} the most sensitive update directions when the new data distribution is far from the post-trained model. Across continual learning settings including (1) distilling from expert traces, (2) self-improvement with STaR and Pedagogical RL, and (3) injecting knowledge after pretraining cutoff, grafting Pareto-dominates both SFT and OPSD in new-task and old-task performance, while avoiding expensive on-policy sampling. Therefore, our work challenges on-policy training as a necessity for continual learning on RL-trained models.

\end{abstract}

%% file: tex/intro.tex
\section{Introduction}
\label{sec:intro}
A long-standing goal of AI is a model that keeps acquiring new capabilities without forgetting what it can already do \citep{mccloskey1989catastrophic,
kirkpatrick2017overcoming}. This is particularly challenging for models that have undergone extensive post-training. In particular, direct supervised finetuning (SFT) on new data can cause forgetting and fail to generalize \citep{chu2025sftmemorizes}. This has motivated the view that continual learning should remain \textit{on-policy}, i.e. the model learns from behavior produced by its current policy \citep{chen2025retaining}. However, extended on-policy RL can plateau and lose coverage after exhausting non-zero reward behaviors to amplify \citep{yue2025does, zhang2026interplay}. Furthermore, data containing new knowledge or capabilities are often \textit{off-policy}, such as new knowledge after the pretraining cutoff \cite{jang2022ckl}, expert
demonstrations \cite{taori2023alpaca}, delayed feedback from environments \cite{trajectory2026scalingsdpo}, and synthetic data for self-improvement \cite{chakraborty2026pedagogicalrl}.

Recent work on on-policy self-distillation (OPSD) tries to bridge this gap by converting off-policy data into on-policy updates, distilling from a solution-conditioned copy of the model \citep{zhao2026selfdistilledreasoner,hubotter2026rlsd,shenfeld2026sdft}. However, conditioning on the solution can suppress exploration, leak privileged information into the teacher's reasoning, and degrade out-of-domain performance or collapse reasoning length \citep{thinking-degradation-a,thinking-degradation-b}. More fundamentally, this raises a simple question:

\begin{tcolorbox}[colback=violet!8, colframe=violet!40, boxrule=0.5pt, arc=2mm]
\begin{center}
\itshape 
Is on-policy training necessary for continual learning, or are we using off-policy updates poorly?
\end{center}
\end{tcolorbox}

We premise with a surprisingly simple finding: model merging, i.e., by interpolating the SFT-ed weights back toward the original post-trained model \cite{wortsman2022wiseft}, $\theta_{\text{merge}} = (1-\lam)\,\thpost + \lam\,\thpost'$ where $\thpost'$ is the model after SFT on a task, SFT is elevated to a strong baseline (Figure~\ref{fig:teaser}, left), achieving higher performance on the new task and comparable performance on old tasks.
Since merging only scales the SFT update down without changing its direction, this demonstrates SFT learns a useful signal from the new data. However, naive integration of SFT updates clearly interferes with the post-trained model's capabilities. This leads to a tangible key objective:

\begin{figure}[t]
\centering
\iftoggle{arxiv}{\includegraphics[width=\linewidth]{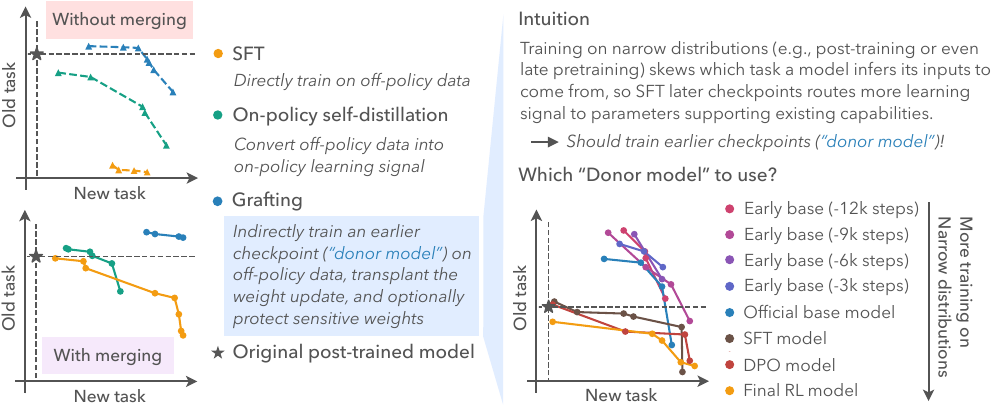}}{\includegraphics[width=\linewidth]{figures/teaser.pdf}}
\caption{\label{fig:teaser} \textbf{(Left) Model merging alone greatly improve generalization; Grafting dominates SFT and OPSD.} Model: Qwen3-1.7B (thinking). \textbf{(Right) The best donor model for grafting is even before the end of \textit{pretraining}.} Model: OLMo-3-7B (thinking). \iftoggle{arxiv}{Only frontier points are plotted.}{}
}
\end{figure}

\begin{tcolorbox}[colback=violet!8, colframe=violet!40, boxrule=0.5pt, arc=2mm]
\begin{center}
    \itshape 
    How can we better minimize the interference induced by off-policy updates?
\end{center}
\end{tcolorbox}
We first rewrite merging as adding a scaled update to the post-trained model, $\theta_{\text{merge}} = \thpost + \lam\,(\textcolor{donorc}{\thpost'} - \textcolor{donorc}{\thpost})$. Written this way, merging decouples two design choices: \emph{where} the update is learned (here, naively on $\thpost$ itself) and \emph{how} it is applied (here, naively scaled by $\lam$).

Task arithmetic \citep{ilharco2023task} shows that an update need not be derived from the model it is applied to. We argue that this separation is important for continual learning. This is because training on a narrow distribution skews which task a model infers its inputs to come from \citep{kotha2024implicit}, so finetuning the post-trained model routes much of the new learning signal through the parameters supporting its existing capabilities, while a checkpoint from before post-training routes less of it there (\S\ref{sec:theory}). We therefore learn the update on an earlier \textcolor{donorc}{donor model} $\textcolor{donorc}{\thdonor}$, and apply it to the post-trained model: $\thpost + \lam\,(\textcolor{donorc}{\thdonorft} - \textcolor{donorc}{\thdonor})$, where $\thdonorft$ is the donor after finetuning. Along the OLMo-3 training trajectory \citep{olmo3}, we see that the best donors are checkpoints before the end of \textit{pretraining} (Figure~\ref{fig:teaser}, right), which is perhaps not surprising since later stages of modern pretraining are high-quality data that resembles post-training.

Donor-trained updates with naive scaling write less to the parameters on which the post-trained model is most sensitive, but when the new distribution is far from the post-trained model, they can still damage them. We therefore propose to optionally protect the most sensitive directions, estimated with the Fisher information \citep{kirkpatrick2017overcoming, ikram2026crispedit}. This matters most for knowledge injection (\S\ref{sec:knowledge}), where retaining unrelated capabilities is hard. We call the full recipe \textbf{grafting}: $\thpost + \lam \cdot \maskrho (\textcolor{donorc}{\thdonorft} - \textcolor{donorc}{\thdonor})$, where the mask $\maskrho$ projects away the top-$\rho$ fraction of Fisher-weighted directions from the update.

We study three settings where off-policy data arises naturally: (1) \textbf{distillation from expert traces} on hard problems the model cannot solve; (2) \textbf{self-improvement} on hard problems, where models bootstrap themselves without stronger supervision, e.g., via their own rejection-sampled successes \citep{zelikman2022star} or with indirect supervision, e.g., from thinking traces written by a copy of themselves with privileged information
\citep[Pedagogical RL;][]{chakraborty2026pedagogicalrl}, and (3) \textbf{knowledge injection} of new facts absent from prior training. In all three, we assume access only to the data (with our without thinking traces depending on the specific setting), not to the weights or logits of a stronger teacher model, so on-policy distillation (OPD) from a teacher \citep{agarwal2023onpolicy} is not possible. 
In all three settings, grafting dominates OPSD in both new-task and old-task performance while avoiding expensive on-policy sampling. 
In self-improvement, grafting roughly doubles new-task performance over the post-trained model while also improving old-task performance. 

Together, our results suggest that on-policy updates are not a necessary condition for continual learning. Since continual learning applications often involve off-policy experiences, we hope our work inspires more effort on making off-policy training generalize.

%% file: tex/preliminaries.tex
\section{Preliminaries}
\label{sec:prelim}

\label{sec:prelim-problem}
We consider teaching a post-trained model $\pi_{\thpost}$ a new target task, formally a set of inputs $\{x\}$ which we want the model to solve. A task may be a skill
(e.g., a class of hard math problems) or some knowledge (e.g., new facts after the knowledge cutoff). Two objectives should be optimized jointly:
\begin{itemize}
[
    noitemsep,
    topsep=-2pt,
    parsep=2pt,
    partopsep=0pt,
    left=15pt
]
  \item \textbf{New task}: performance on held-out IID instances of the target task.
  \item \textbf{Old task}: performance on tasks outside the target task -- for unrelated tasks, we want less forgetting; for related tasks, we want the new skill or knowledge to generalize to these tasks.
\end{itemize}

\label{sec:prelim-onpolicy}
\textbf{Reinforcement learning (RL)} is a standard way to learn a task without using off-policy data. It maximizes
$\mathbb{E}_{y\sim\pi_\theta(\cdot\mid x)}\,R(x,y)$ over
outputs it samples itself. However, given its mode-seeking nature, RL is known to mostly amplify and compose behaviors the model already possesses \citep{zhang2026interplay}. For genuinely new knowledge or very hard problems beyond the model's capability,
the model places almost no probability on correct outputs and, therefore, receives no training signal to continue improving. These are the scenarios where off-policy data can be extremely useful. 

\label{sec:prelim-setting}
Off-policy data can be formalized as a corpus $\mathcal{D}=\{(x,y)\}$ of inputs $x$ and outputs $y$, where $x$ are not generated under $\pi_{\thpost}$.
Such data is abundant in practice, e.g., new knowledge acquired after the model's pretraining cutoff, expert
demonstrations for a hard task, feedback from new environments, and synthetic data created by the model for self-improvement. 

\textbf{Supervised fine-tuning (SFT)} has been the standard way to learn from off-policy data, optimizing the model likelihood $-\mathbb{E}_{(x,y)\sim\mathcal{D}}\log \pi_\theta(y\mid x)$ directly on the outputs $y$ given by the data $\mathcal{D}$. However, SFT is known to memorize rather than generalize \citep{chu2025sftmemorizes} and to cause catastrophic forgetting of capabilities outside $\mathcal{D}$ \citep{mccloskey1989catastrophic, kirkpatrick2017overcoming, biderman2024lora}. 

\textbf{On-policy self-distillation (OPSD)} addresses the shortcomings of SFT by converting the off-policy corpus into on-policy updates \citep{zhao2026selfdistilledreasoner,shenfeld2026sdft,hubotter2026rlsd}. The model samples responses and is supervised at each token by a solution-conditioned version of itself:
\[
\mathbb{E}_{(x,y)\sim\mathcal{D}}\;\mathbb{E}_{\tilde y\sim \pi_\theta(\cdot\mid x,\,y)}\,\sum_{t}
\mathrm{D}_\alpha\!\big(\pi_{\bar\theta}(\cdot\mid x, y, \tilde y_{<t})\,\big\|\,\pi_\theta(\cdot\mid x, \tilde y_{<t})\big),
\]
where the teacher $\bar\theta$ is a stop-grad copy of the student 
, and $\mathrm{D}_\alpha$ is a family of $f$-divergences \citep{agarwal2023onpolicy} whose $\alpha$ interpolates from forward KL ($\alpha{=}0$) to reverse KL ($\alpha{=}1$).
        

%% file: tex/theory.tex
\section{Why does SFT interfere with post-trained capabilities?}
\label{sec:theory}

Why does SFT interfere with post-trained capabilities? We argue that this is because models trained on narrow distributions (e.g., post-training and late stages of pretraining) explain new data differently, also known as \textit{implicit task inference} \cite{kotha2024implicit}. We demonstrate this intuition with a minimal model.

Consider two linear regression tasks with data distributions
$\mathcal{D}_A$ and $\mathcal{D}_B$. For each task $T\in\{A,B\}$,
\begin{equation*}
x \sim \mathcal{D}_T,
\qquad
y = \beta_T^\top x,
\end{equation*}
Task $A$ is a post-trained capability we want to retain, while task $B$ is a capability
we want to continually learn. We model the two tasks with a mixture of linear experts,
\begin{equation*}
\hat y(x)
=
g_\phi(x) w_A^\top x
+
(1-g_\phi(x)) w_B^\top x,
\qquad
g_\phi(x)=\sigma(\phi^\top x).
\end{equation*}
We assume pretraining learns both tasks imperfectly but learns their assignment
relatively well. On task-$B$ inputs, assume for simplicity that $g_{\mathrm{pre}}(x)=\varepsilon_D, x\sim\mathcal{D}_B$. After post-training on task $A$, the gate becomes more biased toward expert $A$ on the same task-$B$ inputs $g_{\mathrm{post}}(x)=\varepsilon_P, x\sim\mathcal{D}_B$, with $0 \leq \varepsilon_D < \varepsilon_P < 1$, which is shown by \citet{kotha2024implicit}.
Thus $\varepsilon$ measures how much task-$B$ data is routed through
expert $A$. 

We now analyze fine-tuning on task $B$. For a checkpoint with task-$B$ assignment $\varepsilon$,
under squared loss,
\begin{align*}
L_B
&=
\frac{1}{2}
r_B^\top \Sigma_B r_B, \quad \text{where} \\
r_B
&=
\varepsilon w_A
+
(1-\varepsilon)w_B
-
\beta_B, \\
\Sigma_B
&=
\mathbb{E}_{x\sim\mathcal{D}_B}[xx^\top].
\end{align*}

The gradients with respect to the two experts are
\begin{equation*}
\nabla_{w_A} L_B
=
\varepsilon \Sigma_B r_B,
\qquad
\nabla_{w_B} L_B
=
(1-\varepsilon)\Sigma_B r_B.
\end{equation*}
Therefore the same task-$B$ learning signal is routed to the two experts in
the ratio
\begin{equation}
\frac{\|\nabla_{w_A}L_B\|}
{\|\nabla_{w_B}L_B\|}
=
\frac{\varepsilon}{1-\varepsilon}.
\end{equation}

Hence task assignment directly controls how much fine-tuning on task $B$
modifies the expert supporting task $A$. Since we assume $\varepsilon_D < \varepsilon_P$,
finetuning the pretrained model routes less of the task-$B$ learning signal
through task $A$ than directly fine-tuning the post-trained
model. This is aligned with what we show in \S\ref{sec:analysis}. 

%% file: tex/method.tex
\section{Method}
\label{sec:method}

\begin{wrapfigure}[7]{r}{0.5\linewidth}
\vspace{-40pt}
\begin{algorithm}[H]
\DontPrintSemicolon

\setlength{\algomargin}{2em}
\caption{Grafting}
\label{alg:graft}
\KwIn{post-trained model $\thpost$, injection data $\mathcal{D}$, scale $\lambda \in (0, 1]$. }
1. $\thdonorft \leftarrow$ finetune $\thdonor$ on $\mathcal{D}$\;
2. $\Delta \leftarrow \thdonorft - \thdonor$\;
3. $\mask_\rho \leftarrow$ compute projection from Eq.~\ref{eq:protect}\;
4. $\theta(\lam, \rho) \leftarrow \thpost + \lam \cdot \mask_\rho(\Delta)$\;
\Return $\theta(\lam, \rho)$
\end{algorithm}
\end{wrapfigure}

Our method \textbf{grafting} consists of two main components: a \textit{scaled} parameter delta derived from training on a \emph{donor model} (i.e., a checkpoint from earlier in training), and an optional \emph{projection mask} to protect sensitive weight directions when the new task data is far from the post-trained distribution. We display the core recipe in Algorithm~\ref{alg:graft}.

\textbf{Extract updates from a donor model \ \ } Task arithmetic \citep{ilharco2023task} shows that an update need not be derived from the model it is applied to. We argue that this separation is important for continual learning. This is because training on a narrow distribution skews which task a model infers its inputs to come from \citep{kotha2024implicit}, so finetuning the post-trained model routes much of the new learning signal through the parameters supporting its existing capabilities, while a checkpoint from before post-training routes less of it there (\S\ref{sec:theory}). Therefore, instead of directly finetuning the post-trained model $\thpost$, we finetune an earlier checkpoint which we call a ``donor model'' $\thdonor$. Specifically, we finetune $\thdonor$ to
obtain $\thdonorft$, and isolate the difference $\Delta = \thdonorft - \thdonor$. We then apply the update scaled by $\lambda$ to $\thpost$:
\begin{equation}
\label{eq:graft}
\theta(\lam) \;=\; \thpost \;+\; \lam\,\big(\thdonorft - \thdonor\big).
\end{equation}
While $\lambda \in (0, 1]$ controls the scale of the graft, we remark that $\lambda$ admits an interpretation that goes beyond interpolating between old and new task performance. This arises from the \emph{model merging} literature (see e.g., \texttt{WiSE-FT}, \citealp{wortsman2022wiseft}), where linearly interpolating a later checkpoint with an earlier one is known to benefit generalization. 
This suggests that moderate $\lambda$ can be beneficial, which is reflected by our experiments on grafting from hard math data. 

\newcommand{\Dold}{\mathcal{D}_{\text{old}}}
\newcommand{\KFAC}{\text{KFAC}\xspace}
\newcommand{\Fret}{F_{\mathrm{ret}}}
\newcommand{\Srho}{S_\rho}
\textbf{Protect sensitive weights via projection \ \ } As we will show, applying a donor-derived update directly may not be enough when the injected distribution is far from the post-trained model (e.g., injecting new facts into a reasoning model). However, precisely characterizing weight interference is complex. We take inspiration from on-policy methods, and consider a surrogate goal of shaping the update direction such that the updated model modifies the behavior on the old domain as little as possible. Concretely, we want an update $\Delta$ that keeps the conditional KL divergence $\E_{z \sim \Dold}\brac{\KL(p_{\thpost}(\cdot \,|\, z) \parallel p_{\thpost+\Delta}(\cdot \,|\, z))}$ over the old domain small. To efficiently translate this to the directions of $\Delta$ to mask, we recall the standard quadratic approximation of KL \citep{cover1991elements}:
\begin{align}\label{eq:KL approx}
    \KL(p_{\thpost}(\cdot \,|\, z) \parallel p_{\thpost+\Delta}(\cdot \,|\, z)) &\approx \frac{1}{2} \Delta^\top F_z(\thpost) \Delta,
\end{align}
where $F_z(\thpost) = \mathbb{E}_{y \sim p_{\thpost}(\cdot\,|\,z)}\!\Big[\big(\nabla_\theta
\log p_{\thpost}(y\,|\,z)\big)^{\otimes 2}\Big]$ is the conditional Fisher Information evaluated at $\thpost$. This can be estimated with a retention set $\mathcal{D}_{\mathrm{ret}} = \{z_1,\dots,z_N\}$ of inputs from, e.g., an old task:
\begin{equation}
\label{eq:fisher}
\Fret :=\; \frac{1}{N}\sum_{n=1}^{N}
\mathbb{E}_{y \sim p_{\thpost}(\cdot\,|\,z_n)}\!\left[\Big(\nabla_\theta
\log p_{\thpost}(y\,|\,z_n)\Big)^{\otimes 2}\right].
\end{equation}
However, the full Fisher is infeasible to compute. We consider existing approaches that take a diagonal (i.e., entrywise) approximation \citep{kirkpatrick2017overcoming}. We also propose another efficient approximation informed by the \emph{Kronecker-Factored Approximate Curvature} (\KFAC) \citep{martens2015optimizing}, similar to approaches in model-editing literature \citep{ikram2026crispedit}, which we can have achieve better performance with less masking than the diagonal approximation. For concreteness, we walk through deriving the mask from the diagonal approximation. Measuring the Fisher quadratic form along the diagonal: $\Delta^\top \diag(\Fret) \Delta = \sum_{j=1}^{d_{\theta}} (\Fret)_{jj} \Delta_j^2$, we observe the entry-wise score  
$s_j = (F_{\mathrm{ret}})_{jj}\,\Delta_j^2$ approximates how much entry $j$ in the update $\Delta$ contributes to the model change (Eq.~\ref{eq:KL approx}). Therefore, a natural approach to reducing the model change induced by $\Delta$ is to \emph{mask} the top $\rho$-fraction of entries $\Srho$ ranked by $s_j$:
\begin{equation}
\label{eq:protect}
\theta(\lam, \rho) \;=\; \thpost \;+\; \lam\,\big(\mask_\rho \odot \Delta \big) ,
\qquad
(\mask_\rho)_j \;=\;
\begin{cases}
0 & j \in \Srho,\\[2pt]
1 & \text{otherwise},
\end{cases}
\end{equation}
where $\odot$ is the element-wise product. We defer analogous derivation of the \KFAC-based projection and other implementation details in this section to \Cref{app:fisher}.
In our experiments, we show that retention-based projections derived from the Diagonal or \KFAC Fisher work remarkably well in removing interference from donor updates by just masking a small number of directions ($\rho \leq 0.01$), outperforming prior Fisher-based merging \citep{matena2022fisher}, and pushing the old/new-domain performance frontier compared to un-projected updates.

\section{Experiments on Learning Hard Math Problems}
\label{sec:math}

\subsection{Experimental setup}
\label{sec:math-setup}

\textbf{Models \ \ } We experiment with two model pairs: Qwen3-1.7B (thinking) / Qwen3-1.7B-Base \cite{qwen3} and
Qwen3-8B (thinking) / Qwen3-8B-Base as the post-trained model $\thpost$ and the donor model
$\thdonor$. Our experiments include both thinking and instruct modes. Unless stated otherwise the
donor model is the officially released base model of the same family.
In \S\ref{sec:olmo} we additionally use OLMo-3-7B (thinking) \citep{olmo3} as $\thpost$, because it
releases intermediate checkpoints along both its pretraining and post-training trajectories, which
lets us vary $\thdonor$ rather than fix it to the official base model.

\textbf{Math data \ \ } We consider three settings for where the off-policy data comes from (see \S\ref{app:data}):
\begin{itemize}[
    noitemsep,
    topsep=-2pt,
    parsep=2pt,
    partopsep=0pt,
    left=15pt]
  \item \textit{Distillation with teacher traces}: we get teacher traces from the AoPS subset of Nemotron-SFT-Math-v4
        \citep{nvidia2026nemotronsftmathv4, du2025nemotronmath}, whose solutions are
        generated by DeepSeek-V4-Pro \citep{deepseekai2026v4}. At both scales we mine hard
        problems from this pool that the post-trained model cannot solve. For each model, we sample $16$ attempts per problem and keep those with $0$ success for 1.7B and $\leq 1$ success for 8B, leaving $1{,}533$ training samples for 1.7B and $985$ for 8B.
  \item \textit{Self-improvement with rejection sampling} (STaR; \citealp{zelikman2022star}): we apply rejection sampling to
        the post-trained model to get supervision of both thinking and solution tokens. We sample up to $32$ attempts per
        problem and keep one verified trace for each solved problem.
  \item \textit{Self-improvement with Pedagogical RL} \citep{chakraborty2026pedagogicalrl}: STaR requires that the model can solve the problem with a large budget. Pedagogical RL removes this requirement by synthesizing reasoning traces from gold solutions and do SFT on the synthesized traces (Appendix~\ref{app:pedrl}). We find replacing the SFT step of Pedagogical RL with grafting largely improves performance.
\end{itemize}
\textbf{Evaluation \ \ } For math, new-task performance is measured on around $100$ held-out hard math problems from the training set. Old-task performance is the mean over three benchmarks (AIME 2024, AIME 2025, HMMT 2025).
All metrics are estimated over $16$ samples at temperature $1.0$ and
top-$p$ $1.0$, decoding up to $32$K tokens in thinking mode, unless otherwise specified.
Appendix~\ref{app:training} gives the training details.

\subsection{Model merging alone improves both old and new tasks}
\label{sec:exp-scaling}

We first show that model merging already improves both old and new tasks. We consider three methods: SFT, OPSD, and grafting, with full details given in Appendix~\ref{app:training}. In Figure~\ref{fig:scaling}, we sweep all methods with $9$ learning rates ($5{\times}10^{-7}$ to $5{\times}10^{-5}$) and train $5$ epochs. Figure~\ref{fig:scaling} (left) shows that, without model merging, neither SFT nor OPSD outperforms the post-trained model, while grafting with the best hyperparameters does. We then sweep the merging scale for all methods in Figure~\ref{fig:scaling} (right). We see that merging improves
generalization and
erases OPSD's original edge over SFT, despite its
slow and expensive on-policy sampling. Given its uniform benefit, we apply
merging across all methods in the rest of the paper.

\begin{figure}[ht]
\centering
\includegraphics[width=0.8\linewidth]{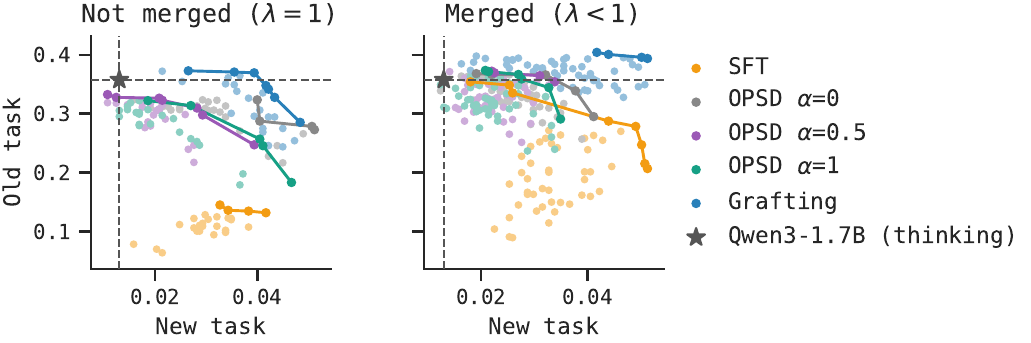}
\caption{\textbf{Model merging improves both old and new task performance.}}
\label{fig:scaling}
\end{figure}

\subsection{Optimal $\thdonor$ is before the end of pretraining}
\label{sec:olmo}

OLMo-3-7B (thinking) \citep{olmo3}
has many checkpoints for us to move $\thdonor$ freely. On the post-training side, we have base model $\xrightarrow{\text{SFT}}$ SFT checkpoint $\xrightarrow{\text{DPO}}$ DPO checkpoint $\xrightarrow{\text{RL}}$ OLMo-3-7B (thinking); on the pretraining side, we move $\thdonor$ up to $12$k steps before the base model.

\begin{figure}[ht]
\centering
\begin{subfigure}[t]{0.52\linewidth}
\centering
\includegraphics[width=\linewidth]{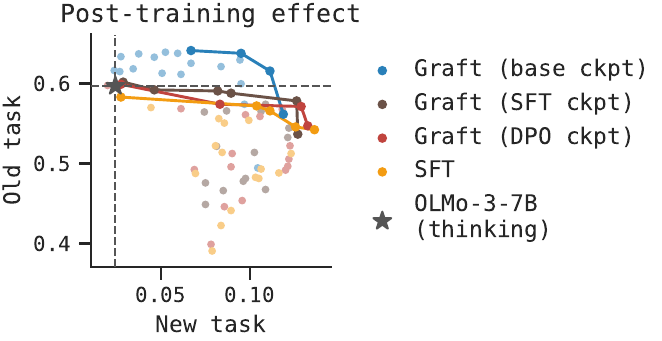}
\end{subfigure}
\hfill
\begin{subfigure}[t]{0.46\linewidth}
\centering
\includegraphics[width=1.0\linewidth]{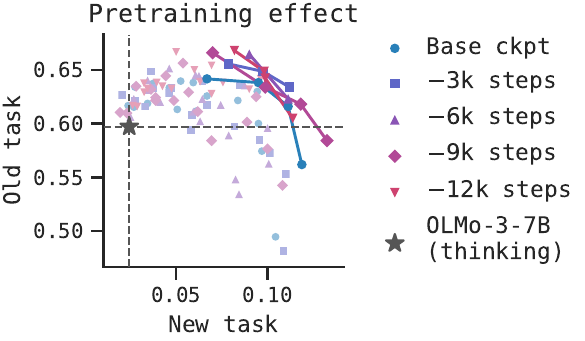}
\end{subfigure}
\caption{\textbf{Grafting from different $\thdonor$ with OLMo-3-7B (thinking).} (Left) $\thdonor$ is
varied along the post-training trajectory: base model $\xrightarrow{\text{SFT}}$ SFT checkpoint $\xrightarrow{\text{DPO}}$ DPO checkpoint $\xrightarrow{\text{RL}}$ OLMo-3-7B (thinking). (Right) $\thdonor$ is moved back along pretraining, up to $12$k steps before the base model.}
\label{fig:olmo}
\end{figure}

We first investigate how each post-training stage makes finetuning harder. We see in Figure~\ref{fig:olmo} (left) that grafting (base ckpt) outperforms grafting (SFT ckpt), grafting (SFT ckpt) outperforms grafting (DPO ckpt), and grafting (DPO ckpt) outperforms direct SFT on the final post-trained model. This suggests that each stage of post-training (SFT, DPO, RL) makes OLMo-3 harder to finetune.

The question that remains is which pretraining checkpoint to use as the donor model. There is a tension in this choice. The end of
pretraining anneals on higher-quality data and already resembles an early stage of SFT, so an earlier $\thdonor$ should be easier to finetune, but it also
sits farther from $\thpost$, which makes it less likely to stay linearly connected
\citep{frankle2020linear, neyshabur2020transfer} with $\thpost$.
Figure~\ref{fig:olmo} (right) moves $\thdonor$ up to $12$k steps before the final base checkpoint. We see that earlier pretrained checkpoints clearly outperforms the final pretrained checkpoint, while the gain is not strictly monotonic. This suggests that for grafting we should search for an early pretrained checkpoint that balances the two forces. 

Qwen3 releases only its final base model and no earlier pretraining checkpoints, so in the rest of the paper we use the official base model as $\thdonor$.

\begin{figure}[ht]
\centering
\begin{subfigure}[t]{\linewidth}
\centering
\includegraphics[width=0.97\linewidth]{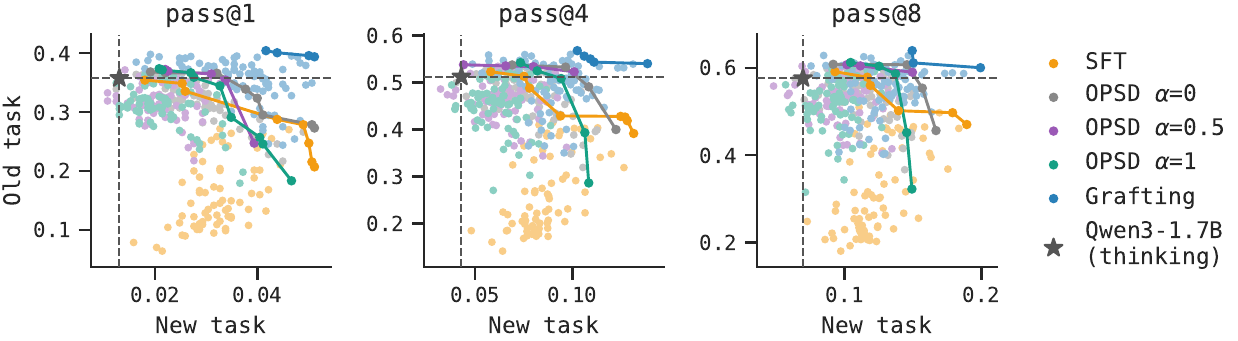}
\caption{Qwen3-1.7B (thinking)}
\label{fig:atlas}
\end{subfigure}
\vspace{5pt}

\begin{subfigure}[t]{0.49\linewidth}
\centering
\includegraphics[width=0.906\linewidth]{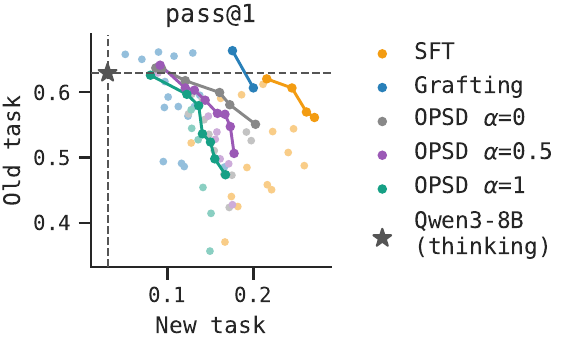}
\caption{Qwen3-8B (thinking)}
\label{fig:atlas8b}
\end{subfigure}\hfill
\begin{subfigure}[t]{0.49\linewidth}
\centering
\includegraphics[width=0.906\linewidth]{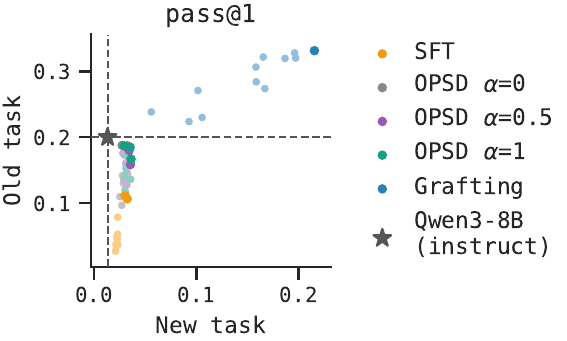}
\caption{Qwen3-8B (instruct)}
\label{fig:atlas-nt}
\end{subfigure}
\caption{\textbf{Distillation from teacher traces}  across different models.}
\label{fig:atlas-all}
\end{figure}

\subsection{Results for distillation with teacher traces}
\label{sec:exp17b}
\textbf{Thinking mode \ \ } Figure~\ref{fig:atlas} and Figure~\ref{fig:atlas8b} show results
with teacher traces as off-policy data, on Qwen3-1.7B (thinking) and Qwen3-8B (thinking), respectively. The same merging sweep (\S\ref{sec:exp-scaling}) is applied for all methods. We see that, for 1.7B, grafting dominates both
SFT and OPSD and is the only method that meaningfully dominates the post-trained model. The same
ordering holds at pass@4 and pass@8, suggesting that the gain reflects expanded capability and
not simply a ``sharpening effect''.
For 8B, both grafting and SFT dominate OPSD. For both sizes, OPSD fails to largely outperform the original post-trained model for generalization, which echoes the findings in \citet{thinking-degradation-a,thinking-degradation-b}.

\textbf{Non-thinking mode (aka. instruct mode) \ \ }
We repeat the experiment at 8B in \emph{non-thinking} mode on DAPO \citep{yu2025dapo}, an
easier data for the non-thinking mode to learn, and we follow the thinking mode experiments to mine hard problems (see Appendix~\ref{app:data} for details). The same merging sweep (\S\ref{sec:exp-scaling}) is applied for all methods. We see from
Figure~\ref{fig:atlas-nt} that grafting again dominates SFT and OPSD at pass@$1$.

\subsection{Self-improvement with rejection sampling (STaR)}
\label{sec:exp-star}
\begin{wrapfigure}[14]{r}{0.47\linewidth}
\vspace{-5pt}
\centering
\includegraphics[width=0.966\linewidth]{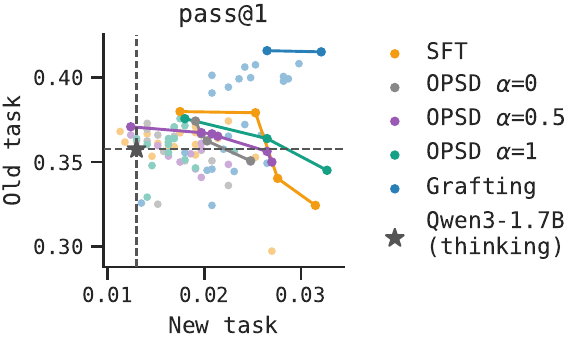}
\caption{\textbf{Self-improvement with rejection \\sampling (STaR)} on Qwen3-1.7B (thinking).}
\label{fig:star-star}
\end{wrapfigure}
We now turn to the second setting, where the off-policy data come from the post-trained model itself via rejection sampling (\S\ref{sec:math-setup}), also known as STaR \citep{zelikman2022star}. Specifically, we sample up to $32$ attempts per problem and keep one verified trace for each solved problem. Unlike distillation from teacher traces, STaR lets the model self-improve by training on its own data. 
Figure~\ref{fig:star-star} shows that new-task performance is roughly the same compared to the strongest baseline (i.e., SFT, which stands for the original STaR) but generalization to old task rises about eight points above it. Compared to distillation from teacher traces (Figure~\ref{fig:atlas}), we see that STaR is generally worse in terms of new-task performance but similar for generalization to the old task.

\subsection{Self-improvement with Pedagogical RL}
\label{sec:exp-pedarl}
\begin{wrapfigure}[15]{r}{0.5\linewidth}
\centering
\includegraphics[width=0.972\linewidth]{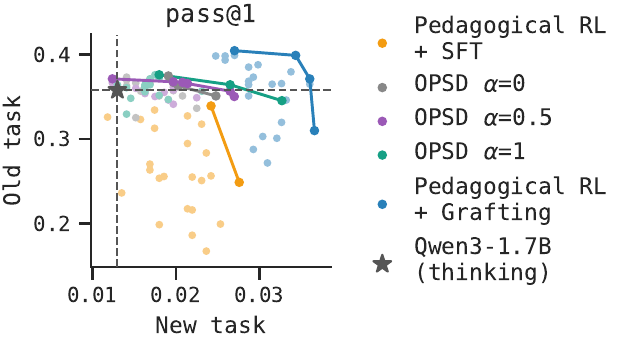}
\caption{\textbf{Self-improvement with Pedagogical RL} on Qwen3-1.7B (thinking).}
\label{fig:star-pedrl}
\end{wrapfigure}
Pedagogical RL~\citep{chakraborty2026pedagogicalrl} takes a different route to self-improvement, where a copy of the model that sees the reference solution (without seeing thinking traces) is trained with RL to write thinking traces the model can learn from (Appendix~\ref{app:pedrl}). With these self-generated thinking traces, the original method then learns with SFT (Pedagogical RL + SFT in Figure~\ref{fig:star-pedrl}). Compared to direct distillation from teacher traces, Pedagogical RL does not need access to thinking traces, which is more practical since solutions from humans or expert systems do not expose thinking traces. 
We keep the RL training and thinking trace generation steps intact and propose to learn the same traces with grafting. Since Pedagogical RL is designed to generate thinking traces and OPSD does not use them, we reuse the OPSD results from STaR (\S\ref{sec:exp-star}). Figure~\ref{fig:star-pedrl} shows that grafting again dominates SFT and OPSD. With Pedagogical RL, grafting learns the new task better and generalizes to the old task, while SFT loses in both directions.

\section{Experiments on Knowledge Injection}

\subsection{Experimental setup}

\textbf{Models \ \ } We experiment with Qwen3-1.7B (thinking) as the post-trained model $\thpost$ and Qwen3-1.7B-Base \cite{qwen3} as the donor model $\thdonor$. 

\textbf{Knowledge data \ \ } Our data for knowledge acquisition mainly follows \citet{shenfeld2026sdft}, but with the key exception that we use purely synthetic knowledge to avoid any possible leakage. We construct a fictional world with $200$ invented entities and $1{,}125$
atomic facts. Each fact is written as question-answer
pairs in $8$ distinct phrasings, and the model is trained on those; evaluation then uses
\emph{held-out paraphrases} the model has never seen. For OPSD, we write wiki-style articles that state the fictional facts for conditioning. Since the knowledge data's QA form is very different from a reasoning model's, we mix knowledge QA with AoPS teacher
traces we use in the math experiments. Together, we have $9{,}000$ QA samples and $1{,}533$ AoPS
traces. All methods are trained on this same mixture.

\textbf{Evaluation \ \ } For knowledge, new-task performance is the exact-match accuracy on held-out paraphrases the model never saw, decoded greedily. Old-task performance is the mean over three benchmarks (AIME 2024, AIME 2025, HMMT 2025). Appendix~\ref{app:training} gives the training details.

\begin{figure}[b]
\centering
\begin{subfigure}[t]{0.46\linewidth}
\centering
\includegraphics[width=0.965\linewidth]{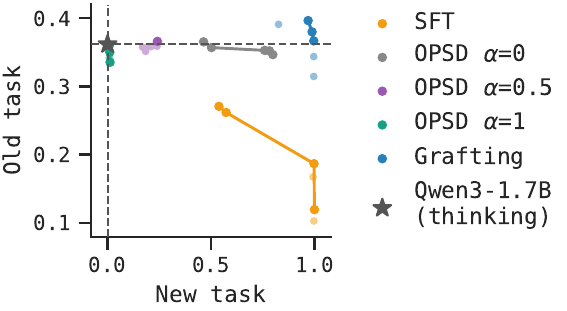}
\end{subfigure}%
\hfill
\begin{subfigure}[t]{0.51\linewidth}
\centering
\includegraphics[width=0.976\linewidth]{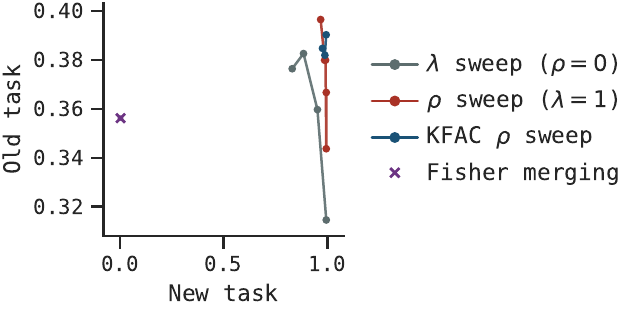}
\end{subfigure}
\caption{\textbf{(Left) Knowledge injection results.} \textbf{(Right) Disentangling the contribution of grafting's components.} The protection mask ($\rho$ sweep) dominates no mask ($\lambda$ sweep).}
\label{fig:knowledge}
\label{fig:protect}
\end{figure}

\subsection{Knowledge injection results}
\label{sec:knowledge}

Figure~\ref{fig:knowledge} (left) shows the knowledge injection result with Qwen3-1.7B (thinking), where the same model merging sweep (\S\ref{sec:exp-scaling}) is applied to all methods. We see that grafting
achieves nearly perfect injection with retention at or above the post-trained model. SFT injects just as well but suffers from forgetting. OPSD achieves near-perfect retention but also learns less. Within OPSD, forward KL ($\alpha{=}0$) learns much more than JSD ($\alpha{=}0.5$) and reverse KL ($\alpha{=}1$). This shows that learning a new fact requires a mode-covering objective rather than mode-seeking. 

\section{Analysis} 
\label{sec:analysis}

\textbf{Ablation studies \ \ }
Here we ablate the mask step that zeroes the most sensitive coordinates. Figure~\ref{fig:protect} (right) compares the protection mask ($\rho$ sweep, in red) with plain model merging without the mask ($\lambda$ sweep, in gray). We see that the protection mask Pareto-dominates plain model merging. Figure~\ref{fig:protect} (right) also shows Fisher merging \citep{matena2022fisher}, which weights each coordinate by its Fisher ratio between the two tasks. We see that Fisher merging does not inject at all in this setting. We find that this is because the knowledge injection Fisher is usually dominated by the retention Fisher. Finally, the \KFAC mask (\S\ref{sec:method}, in navy) reaches the same old-task performance as the diagonal mask while zeroing far fewer directions: at $\rho{=}0.001$ it keeps new-task performance at $0.995$ with old-task performance $0.390$, which the diagonal mask needs $\rho{=}0.01$ to match. We note that \KFAC being a better approximant to the retention Fisher does \textbf{not} directly imply interference must be more concentrated to few directions. The fact that retention of old tasks and injected knowledge can be achieved by zero-ing very few directions is in line with parallel work demonstrating the top few directions of curvature contain mostly interference \citep{zhu2026sft}, and can be removed with little effect on task performance \citep{song2025does}.

\begin{figure}[t]
\centering
\begin{subfigure}[t]{0.49\linewidth}
\centering
\includegraphics[width=0.95\linewidth]{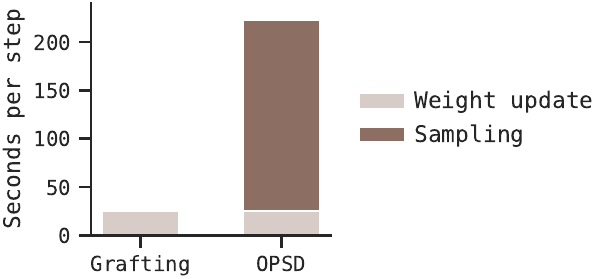}
\end{subfigure}
\hfill
\begin{subfigure}[t]{0.49\linewidth}
\centering
\includegraphics[width=0.88\linewidth]{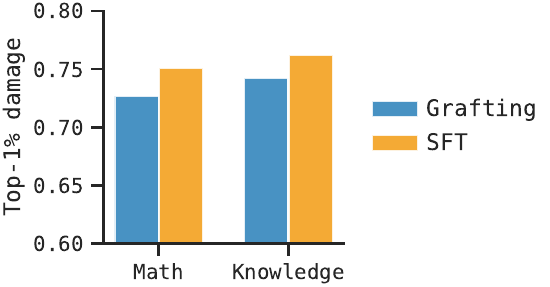}
\end{subfigure}
\caption{\textbf{(Left) Grafting is much faster to train than OPSD.} We show the per-step sampling and weight update time for Qwen3-8B (thinking). \textbf{(Right) Grafting writes less to sensitive coordinates than SFT.} We show the weighted delta $\sum_j F_{\mathrm{ret}}(j)\Delta_j^2$ summed across the $1\%$ of coordinates with the largest $F_{\mathrm{ret}}$.}
\label{fig:cost-mechanism}
\end{figure}

\textbf{Training latency \ \ } Off-policy training also gets rid of the latency cost of on-policy sampling. We show in Figure~\ref{fig:cost-mechanism} (left) the per-step wall clock time for both sampling and weight update for grafting and OPSD, both with Qwen3-8B (thinking). We see that sampling is roughly $90\%$ of the step. In other words, grafting is $10\times$ faster to train than OPSD, while achieving much better performance. 

\textbf{Training on donor model writes less to sensitive weights \ \ } We show in Figure~\ref{fig:cost-mechanism} (right) the Fisher-weighted weight delta $\sum_j F_{\mathrm{ret}}(j)\Delta_j^2$ that falls
inside the top $1\%$ of coordinates with the largest $F_{\mathrm{ret}}$, normalized by that of all coordinates. In other words, this number shows how much a weight update writes into the most sensitive coordinates of the post-trained weights. We first see that the $1\%$ coordinates explain nearly $75\%$ of the weight changes, which empirically justifies the protection mask. More importantly, we see that grafting writes less to the sensitive weights than SFT, for both math and knowledge learning tasks, which helps explain why the protection mask alone cannot push SFT to the same frontier as grafting, as we see above. In \Cref{sec:theory}, we show how this effect can be predicted by a minimal linear model.

%% file: tex/related_work.tex
\section{Related Work}
\label{sec:related}

\textbf{Continual learning for post-trained models. \ \ \ }
Continually trained networks can overwrite earlier capabilities \citep{mccloskey1989catastrophic}.
Classic remedies constrain the update written into the parameters, e.g., elastic weight
consolidation \citep{kirkpatrick2017overcoming}. For language models, it has been shown that low-rank updates \citep{hu2022lora} forget less but also learn less \citep{biderman2024lora}. SFT and RL learn in different ways \citep{chu2025sftmemorizes, matsutani2025rlsqueezes}, and SFT in
particular memorizes at the cost of generalization. This echoes the common belief is that continual training should be on-policy, and has been demonstrated by the success of continual training models using RL with verifiable rewards (RLVR) \citep{deepseekai2025r1}. \citet{chen2025retaining} attribute RL's smaller forgetting specifically to its use of on-policy data, rather than to other algorithmic choices such as KL regularization or advantage estimation. In practice, however, most data
for continual learning is off-policy, such as knowledge after the pretraining cutoff
\citep{jang2022ckl}, expert demonstrations \citep{taori2023alpaca}, delayed feedback from
environments \citep{trajectory2026scalingsdpo}, and synthetic data for self-improvement \cite{zelikman2022star}. 

\textbf{On-policy self-distillation and its failure modes. \ \ \ }
OPSD converts off-policy data into an on-policy learning signal, by distilling a
solution-conditioned copy of the model into itself on its own samples \citep{zhao2026selfdistilledreasoner,shenfeld2026sdft,hubotter2026rlsd,penaloza2026pid}. The approach is
being scaled in practice \citep{trajectory2026scalingsdpo}. Its benefit, however, appears to hold
only within a limited scope of tasks. Reported failure modes include suppressed exploration, degeneration
in reasoning length, and failure to generalize \citep{thinking-degradation-a,
thinking-degradation-b}. Recent work responds to these failures by filtering the
part of the teacher signal that does not transfer to the student
\citep{shen2026purifiedopsd} or by routing token-level supervision between privileged teacher and
student policies \citep{yu2026dopd}.

\textbf{Model merging and weight-space operations. \ \ }
Weight-space operations are promised by loss-landscape geometry \citep{frankle2020linear,
neyshabur2020transfer, ainsworth2023gitrebasin}, and a large literature does weight interpolation with models trained or finetuned with the same initialization
\citep{wortsman2022modelsoups, matena2022fisher, li2022branchtrainmerge, wortsman2022wiseft,
ilharco2022patching}, with refinements that resolve interference, linearize
fine-tuning, or localize a skill in a sparse subset of the delta \citep{yadav2023ties, yu2024dare,
ortizjimenez2023tangent, panigrahi2023skill}. Model merging has been shown to improve robustness \citep{wortsman2022wiseft} and generation diversity \citep{dang2025weightensembling}. Most similar to our work, task arithmetic \cite{ilharco2023task}, also known as Shadow-FT \citep{wu2025shadowft} or chat vector \cite{huang2024chatvector}, composes weight deltas trained on the base model. We improve over them with the proposed Fisher masking and donor model selection. Fisher merging \citep{matena2022fisher} derives a soft weighting between multiple tasks that train the model, while we derive a sparse mask using only the retention data and show that retention-only sparse masking is substantially more effective.

\textbf{Models become harder to finetune. \ \ }
It has been known that training changes how well a model takes a further update, or plasticity \citep{ash2020warmstarting, dohare2024plasticity}. Recent work demonstrates this for large language models. The longer a base model is
pretrained beyond a certain point, the more sensitive its parameters become and the worse it finetunes
\citep{springer2025overtrained}. To mitigate such overtraining effect, it has been proposed to increase learning rate \citep{rofin2026learningrates}, or more broadly, to bias pretraining
toward flatter minima \citep{watts2026sharpnesspretraining}. Our experiments on varying the donor model for grafting point the same
way: an earlier pretraining checkpoint can be a better $\thdonor$ than the final base
model and later post-trained intermediate checkpoints. Different from previous work, we show that the harder-to-finetune effect can be mitigated not only by training an easier-to-finetune model, but also by better continual update rules.

%% file: tex/conclusion.tex
\section{Conclusion and Future Work}
\label{sec:conclusion}
We asked whether continual learning for RL-trained models has to be on-policy, and our results suggest
that it does not. Model merging alone shows that
SFT learns a useful signal from new data, but naively applying its update interferes with existing capabilities. Building on that, grafting reduces this interference by
learning the update on an earlier donor checkpoint, scaling it, and optionally masking the most sensitive update directions. Across three settings, grafting Pareto-dominates both SFT and OPSD in
new-task and old-task performance, while avoiding expensive on-policy sampling. We show that w donor model to use for grafting turns out to be crucial too. Earlier pretraining checkpoints are usually a better choice, though future work can study more principled approaches to select the optimal pretraining checkpoint for grafting. Another direction to optimize for future work is to predict optimal $\lambda$ and $\rho$ instead of a sweep. More broadly, since off-policy data is what continual learning
mostly has, we hope this work encourages more effort on making off-policy training generalize.

%% file: tex/acks.tex
\section*{Acknowledgments} 
We wish to thank Jacob Springer, 
Gaurav Ghosal, 
Ruslan Salakhutdinov,
Weichen Yu,
Neil Kale, 
Jingchu Gai, Christina Baek,
Ziqian Zhong, Kevin Wang for their discussion and feedback. 
This research is partially supported by
the DARPA expMath program through the DARPA CMO contract number HR0011262E028. We gratefully acknowledge support from Schmidt Sciences, NSF, Apple, Open Philanthropy, Google.
We would like to thank CMU FLAME Cluster for providing generous compute.

%% file: appdx/data.tex
\section{Data}
\label{app:data}

\textbf{Knowledge (synthetic closed world). \ \ } To guarantee the facts are new, we build a closed
fictional world rather than reuse public corpora: $200$ invented entities and $1{,}125$ atomic facts
about them. Each fact is written as question-answer pairs in $8$ distinct phrasings, giving $9{,}000$
QA pairs, and new-task performance is measured on held-out phrasings never seen in training, so recall cannot be
inflated by memorized surface form. For OPSD we additionally write $200$ wiki-style articles that
state the facts, used only for conditioning. The facts are absent from the post-trained model:
closed-book exact-match is $\approx\!0\%$ (instruct $0.09\%$, base $0.13\%$). Answers are scored by
rule-based \texttt{<answer>} exact-match. All methods train on the same mixture of the
$9{,}000$ QA pairs and the $1{,}533$ AoPS traces of the math corpus.

\textbf{Math reasoning, Qwen3-1.7B (thinking). \ \ } Training traces are competition math
solutions sampled from the AoPS split of Nemotron-SFT-Math \citep{nvidia2026nemotronsftmathv4}. We mine the hard
subset by sampling Qwen3-1.7B (thinking) $16$ times per problem at
temperature $1.0$ and keeping only problems it fails on all $16$ attempts, so the
target skill is absent. After length-filtering and deduplication we keep $1{,}533$ training
traces, and hold out a disjoint set of $111$ mined-hard problems to measure new-task performance.

\textbf{Math reasoning, Qwen3-8B (thinking). \ \ } At 8B we mine the same pool with a looser criterion (so we have a meaningful number of samples), keeping
problems the post-trained model solves at most once in $16$ attempts, which leaves $985$ training
samples. A disjoint set of $100$ problems is held out to measure new-task performance.

\textbf{Math reasoning, Qwen3-8B (non-thinking). \ \ } The non-thinking experiment uses problems from
DAPO \cite{yu2025dapo}. Hard problems are mined the same way as above, by sampling Qwen3-8B in
non-thinking mode at temperature $1.0$ and keeping only the problems it never solves, and the worked
solutions for those problems are the off-policy data. This leaves $1{,}163$ training samples, with a
disjoint set of $200$ problems held out to measure new-task performance.

%% file: appdx/pedagogical_RL.tex
\section{Pedagogical RL}
\label{app:pedrl}

Pedagogical RL \citep{chakraborty2026pedagogicalrl} improves a model in two stages: (1) it synthesizes
reasoning traces from problems and their reference solutions, and (2) it finetunes the model on the
synthesized traces with SFT. We keep stage (1) unchanged and replace SFT in stage (2) with grafting.

\textbf{Notation \ \ } Following \citet{chakraborty2026pedagogicalrl}, let $x$ denote a problem, $c$ its
reference solution without reasoning, and $\tau$ the reasoning trace to be synthesized. Two copies
of the model play different roles: a privileged teacher
\[
\pi_{\theta_T}(\tau \mid x,c),
\]
whose prompt contains the reference solution and instructs it to solve the problem in its own way
rather than copy $c$, and a student
\[
\pi_{\theta_S}(\tau \mid x),
\]
which sees only the problem.

\textbf{Teacher training \ \ } The teacher is trained to write traces that are both correct and
plausible under the student. For each position $t$ of a trace, let the student's most likely next
token be
\[
a_t^{\max}
=
\arg\max_a
\pi_{\theta_S}
\left(
a \mid x,\tau_{<t}
\right),
\]
and define the surprise gap
\[
d_t
=
\log
\frac{
\pi_{\theta_S}
\left(
a_t^{\max}
\mid x,\tau_{<t}
\right)
}{
\pi_{\theta_S}
\left(
\tau_t
\mid x,\tau_{<t}
\right)
},
\]
which is zero when the trace follows the student's top choice and grows as the trace takes tokens
the student finds unlikely. The spike-aware learnability score aggregates the gaps over the $T$
tokens of the trace,
\[
G_{\mathrm{spike}}^{\theta_S}(\tau \mid x)
=
\exp
\left[
-\frac{\lambda}{\beta}
\log
\left(
\frac{1}{T}
\sum_{t=1}^{T}
\exp(\beta d_t)
\right)
\right],
\]
where $\beta$ interpolates between penalizing the mean gap ($\beta \to 0$) and the largest gap
($\beta \to \infty$), and $\lambda$ sets the strength of the penalty. The pedagogical reward
multiplies the learnability score with the correctness reward $R$,
\[
r_{\mathrm{ped}}(x,c,\tau)
=
R(x,c,\tau)
\,
G_{\mathrm{spike}}^{\theta_S}(\tau \mid x),
\]
and the teacher is trained with GRPO \citep{shao2024deepseekmath} to maximize
\[
\max_{\theta_T}
\;
\mathbb{E}_{
\tau \sim
\pi_{\theta_T}(\cdot \mid x,c)
}
\left[
r_{\mathrm{ped}}(x,c,\tau)
\right].
\]
The student $\pi_{\theta_S}$ stays fixed during teacher training and only scores how learnable the
teacher's traces are. We denote the trained teacher by $\tilde{\pi}_T$.

\textbf{Trace synthesis \ \ } For each problem and its reference solution, we sample reasoning traces
\[
\tau
\sim
\tilde{\pi}_T(\cdot \mid x,c)
\]
from the trained teacher. These traces are the training data for stage (2).

\textbf{Problems and solutions \ \ } The pairs $\{(x, c)\}$ are the $1{,}533$ training samples of the 1.7B teacher-trace setting (\S\ref{sec:math-setup}), covering $628$ unique problems: $x$ is the problem and $c$ is the written solution of the teacher trace with its reasoning removed. Each pair also carries the problem's final answer, which the correctness reward $R$ checks. The synthesized pairs $\{(x, \tau)\}$ are the off-policy data for stage (2), where grafting replaces SFT.

\textbf{Hyperparameters \ \ } The teacher and the student are both initialized from Qwen3-1.7B in
thinking mode, and the maximum sequence length is $18$K tokens.

%% file: appdx/training.tex
\section{Training}
\label{app:training}

\textbf{Grafting and direct SFT (math). \ \ } Both fine-tune with identical optimization and differ
only in the checkpoint updated (the earlier checkpoint for grafting, the post-trained model for direct
SFT). The grid spans nine learning rates ($5{\times}10^{-7}$ to $5{\times}10^{-5}$) and five
epochs, snapshotting after each epoch, at global batch $32$ and sequence length $18$K.
We optimize with AdamW ($\beta{=}(0.9, 0.95)$, weight decay $0.01$) under a cosine schedule with
warmup ratio $0.1$ and gradient clipping at $1.0$.

\textbf{Grafting and direct SFT (knowledge). \ \ } Full fine-tuning at learning rate
$2{\times}10^{-5}$ with a cosine schedule, warmup ratio $0.1$, $\beta=(0.9, 0.95)$, weight decay
$0.01$, gradient clip $1.0$, global batch $32$, and sequence length $18$K. The knowledge QA samples receive an empty-think prefix (\texttt{<think>}\textbackslash n\textbackslash
n\texttt{</think>}), while the replay samples receive the standard \texttt{<think>reasoning\textunderscore content</think>}.

\textbf{On-policy self-distillation (OPSD). \ \ } OPSD follows grafting and SFT in terms of training hyperparameters. Here are OPSD-specific details: the model samples its
own response conditioned on a demonstration and is trained to match a frozen, demonstration-conditioned teacher under the divergence $\mathrm{D}_\alpha$, with rollout $n{=}1$ and top-$k$ $100$ for the teacher logits. We report $\alpha\in\{0,0.5,1\}$ for OPSD.

%% file: appdx/evaluation.tex
\section{Evaluation}
\label{app:eval}
\textbf{Decoding. \ \ } All math scores are averaged over $16$ rollouts at temperature
$1.0$ and top-$p$ $1.0$, decoding up to $32$K tokens in thinking mode and $16$K in non-thinking mode.
Knowledge injection is decoded greedily.

\textbf{New task. \ \ } New-task performance is measured on a held-out set the model was never trained on, as defined in \S\ref{app:data}.

\textbf{Old task. \ \ } Old-task performance is the mean over the same three out-of-domain
competition-math suites, AIME 2024, AIME 2025 and HMMT 2025. The quantity measures generalization
when the continual learning task is math and retention when it is knowledge.

%% file: appdx/olmo3_exps.tex
\section{OLMo-3 experiments}
\label{app:olmo}

\textbf{Data. \ \ } We mine hard problems for OLMo-3-7B (thinking) the same way as for Qwen3-1.7B (thinking):
sampling OLMo-3-7B (thinking) $16$ times per problem at temperature $1.0$ and keeping only
problems it fails on all $16$ attempts. We hold out $100$ problems i.i.d.\ to measure new-task performance, and train on
every trace of the rest.

\textbf{Checkpoints. \ \ } On the post-training side we use the four checkpoints OLMo-3 releases: the
base model, the SFT checkpoint, the DPO checkpoint, and OLMo-3-7B (thinking) itself. On the
pretraining side we use stage-3 (see \citet{olmo3} for details) steps $11{,}921$, which is the released base model, $9{,}000$,
$6{,}000$ and $3{,}000$, together with stage-2 step $47{,}684$, the last stage-2 checkpoint. The gaps
from the released base model are $2{,}921$, $5{,}921$, $8{,}921$ and $11{,}921$ steps, which we write
as $3$k, $6$k, $9$k and $12$k.

\textbf{Training. \ \ } We sweep over five learning
rates ($5{\times}10^{-6}$ to $5{\times}10^{-5}$), for five epochs at global batch $32$ and maximum
length $18{,}432$, in thinking mode.

\textbf{Training stage-2 checkpoints on long outputs. \ \ } Stage-2 base checkpoints \cite{olmo3} set the max lengths as $8{,}192$ with no RoPE
scaling, so we copy stage-3's YaRN block onto them to finetune at $18{,}432$ tokens to enable training stage-2 base checkpoints on long reasoning traces. 

%% file: appdx/fisher.tex
\section{Fisher estimation}
\label{app:fisher}

We discuss the various derivation and implementation details of the retention Fisher \eqref{eq:fisher}. Recall that the Diagonal Fisher approximation leads to scores that weight the donor update by entry, and the resulting projection \eqref{eq:protect} reduces to a $\{0,1\}$ mask corresponding to the entries with the highest scores. In general, let us consider an orthonormal basis $U \in \R^{d_\theta \times d_\theta}$. As examples, the Diagonal Fisher approximation corresponds to the standard coordinate basis $U = [e_i]$. On the other hand, if $U$ is the eigenbasis of $\Fret = U \Sigma U^\top$, $\Sigma = \diag(\sigma_i)$ then we note the corresponding scores and projection:
\begin{align*}
    \Delta^\top \Fret \Delta = (U^\top \Delta)^\top \Sigma (U^\top \Delta) = \sum_{i=1}^{d_\theta} \sigma_i \ip{u_i, \Delta}^2.
\end{align*}
Thus, defining the score accordingly $s_i := \sigma_i \ip{u_i, \Delta}^2$ and the set $S_\rho := \{u_i: s_i \text{ in top-}\rho \text{ fraction}\}$, a mask in the diagonal case turns into a \emph{projection} onto the orthogonal complement of $S_\rho$: $\maskrho := \mathrm{Proj}^\perp_{S_\rho}(\cdot)$. However, we note that the projection based on $S_\rho$ is exact only when $U$ diagonalizes $\Fret$, as otherwise projecting based on scores $s_i$ \textbf{ignores the contribution from off-diagonal elements} $(\Fret)_{ij} \Delta_i \Delta_j$, which can be greatly suboptimal if $\Delta_i$ and $\Delta_j$ are tightly coupled, e.g. parameters in the same weight matrix. The coordinate basis is often a suboptimal choice for this purpose, and an exact eigendecomposition of the full-dimensional Fisher is infeasible.

\newcommand{\din}{d_{\mathrm{in}}}
\newcommand{\dout}{d_{\mathrm{out}}}
\newcommand{\vectorize}{\mathrm{vec}}
\newcommand{\tilDeltaW}{\tilde{\Delta}^W}
This is a similar motivation as prior optimization literature that seeks to approximate Natural Gradient Descent for parametric maximum likelihood estimation, which uses the Fisher Information matrix as a preconditioner. A seminal work on this front is the \KFAC optimizer \citep{martens2015optimizing}. The essential observation is that the Fisher Information matrix can be approximated by a block-diagonal matrix, where each block corresponds to a layer's parameters; then, for matrix-shaped weights $W \in \R^{\dout \times \din}$ this can further be approximated by a \textit{Kronecker factor} between two symmetric matrices of shape $P \in \R^{\dout \times \dout}$, $Q \in \R^{\din \times \din}$. For example, given a feedforward weight matrix $h = Wa$, we have
\begin{align}
    P = \E_z\E_{y \sim p_{\theta}(\cdot\,|\,z)}\brac{\frac{\partial L(\theta; y,z)}{\partial h}\frac{\partial L(\theta; y,z)}{\partial h}^\top}, \quad Q = \E_z[aa^\top],
\end{align}
where $L(\theta; y,z) = -\log p_\theta(y | z)$ is the negative log-likelihood of the model's own generation $y$ given input $z$. We note that for $<2$-D parameters the original Diagonal Fisher approximation is made and scores computed as described in \Cref{sec:method}. Consider the eigendecompositions $P = U \Sigma U^\top$, $Q = V D V^\top$, and for a given update $\Delta$, consider the (appropriately re-shaped) subset corresponding to the current layer $\Delta^{W}$. Observe by the Kronecker mixed-product property we have:
\begin{align*}
    \vectorize(\Delta^W)^\top (Q \otimes P)\vectorize(\Delta^W) &= \Tr(P \Delta^W Q \Delta^W) \\
    &= \Tr(U \Sigma U^\top \Delta^W V D V^\top \Delta^W) \\
    &= \Tr(\Sigma \tilDeltaW D (\tilDeltaW)^\top), \quad \tilDeltaW := U^\top \Delta^W V \\
    &= \sum_{i,j=1}^{\dout, \din}  \sigma_i d_j (\tilDeltaW_{i,j})^2.
\end{align*}
Therefore, for this $\dout \cdot \din$ parameter subset of $\Delta$, the scores become $s_{(i,j)} = \sigma_i d_j (u_i^\top \Delta^W v_j)^2$. Tallying the scores up across all scalar/vector parameters and tensor parameters, the \KFAC-based projection mask thus becomes a projection onto the orthogonal complement of $\{e_k\} \cup \{u_i \otimes v_j\}$, where $s_k, s_{(i,j)}$ belong to the top-$\rho$ of scores across scalar/vector and matrix parameters.

Though a KFAC decomposition does not in general perfectly diagonalize the full Fisher Information, as it does not capture block-to-block dependencies, it is known empirically to be a significantly better decomposition than the diagonal \citep{george2018fast}.

\confpar{Implementation.} To estimate the retention Fisher matrix, we compute the per-token-normalized from $1858$ rollouts of the Qwen3-1.7B thinking model $\thpost$ at temperature $1.0$ prompted with AoPS-style math problems. The $\rho$ score threshold is estimated by taking a $\rho$-quantile of a subsample of the list of scores and using that as the selection threshold.

The Diagonal Fisher projection is implemented efficiently via a mask. The KFAC projection is slightly more arithmetic-intensive to compute due to the layer-wise eigendecompositions and resulting matrix multiplications, but remains efficient due to the layer-wise factorization.
